\documentclass[]{spie}  

\usepackage{subcaption}
\usepackage{setspace}

\usepackage{amsmath,amsfonts,amssymb}
\usepackage{graphicx}
\usepackage[colorlinks=true, allcolors=blue]{hyperref}
\usepackage[leftcaption]{sidecap}

\title{Cross Modal Global Local Representation Learning from Radiology Reports and X-Ray Chest Images}

\author[a]{Nathan Hadjiyski$^*$}
\author[b]{Ali Vosoughi$^*$}
\author[b,c,d,e]{Axel Wismüller}

\affil[a]{Department of Computer Science, University of Rochester, NY, USA}
\affil[b]{Department of Electrical and Computer Engineering, University of Rochester, NY, USA}
\affil[c]{ Department of Imaging Sciences, University of Rochester, NY, USA}
\affil[d]{Department of Biomedical Engineering, University of Rochester, NY, USA}
\affil[e]{Faculty of Medicine and Institute of Clinical Radiology, Ludwig Maximilian University,
Munich, Germany}
\affil[*]{Authors with equal contributions.}

\authorinfo{Further author information: (Send correspondence to Ali Vosoughi)\\Ali Vosoughi: E-mail: mvosough@ece.rochester.edu}

\begin{document} 
\maketitle
\begin{abstract}
Deep learning models can be applied successfully in real-work problems; however, training most of these models requires massive data. Recent methods use language and vision, but unfortunately, they rely on datasets that are not usually publicly available. Here we pave the way for further research in the multimodal language-vision domain for radiology. In this paper, we train a representation learning method that uses local and global representations of the language and vision through an attention mechanism and based on the publicly available Indiana University Radiology Report (IU-RR) dataset. Furthermore, we use the learned representations to diagnose five lung pathologies: atelectasis, cardiomegaly, edema, pleural effusion, and consolidation. Finally, we use both supervised and zero-shot classifications to extensively analyze the performance of the representation learning on the IU-RR dataset. Average Area Under the Curve (AUC) is used to evaluate the accuracy of the classifiers for classifying the five lung pathologies. The average AUC for classifying the five lung pathologies on the IU-RR test set ranged from 0.85 to 0.87 using the different training datasets, namely CheXpert and CheXphoto. These results compare favorably to other studies using UI-RR. Extensive experiments confirm consistent results for classifying lung pathologies using the multimodal global local representations of language and vision information.
\end{abstract}

\keywords{Deep learning, cross-modal learning, computer vision, natural language processing, zero-shot classification, chest X-ray, radiology reports}

\section{Introduction}
Deep learning is prevalent thanks to its success in numerous real-life domains and surpassing classic techniques in performance, cost, and ease of applicability. However, despite the computational efficiency of deep learning models, it requires a massive amount of data, which is scarce and expensive, especially in radiology.
Multimodal learning recently got considerable attention. Language is a widely available form of data in radiology besides images. 
Therefore, one can design deep learning models that use language and vision in different ways to improve the quality of medical image analysis, for instance, by using radiology reports to enhance the performance of downstream tasks through x-ray images. 

This paper addresses multimodal representation learning from a combination of radiology reports and their corresponding medical images. We use x-ray images and radiology reports of a publicly available dataset and tailor it to learn a multimodal representation that encompasses local and global language and visual information. Although many recent methods consider language and vision in multimodal learning, one major problem in radiology's vision and language models is the lack of sufficient datasets, which is well addressed in this paper. In many instances, the codes are publicly available without corresponding datasets, preventing further developments of the methods in the community [\citeonline{huang2021gloria}]. 

\begin{figure*}
    \centering
    \includegraphics[width=\textwidth]{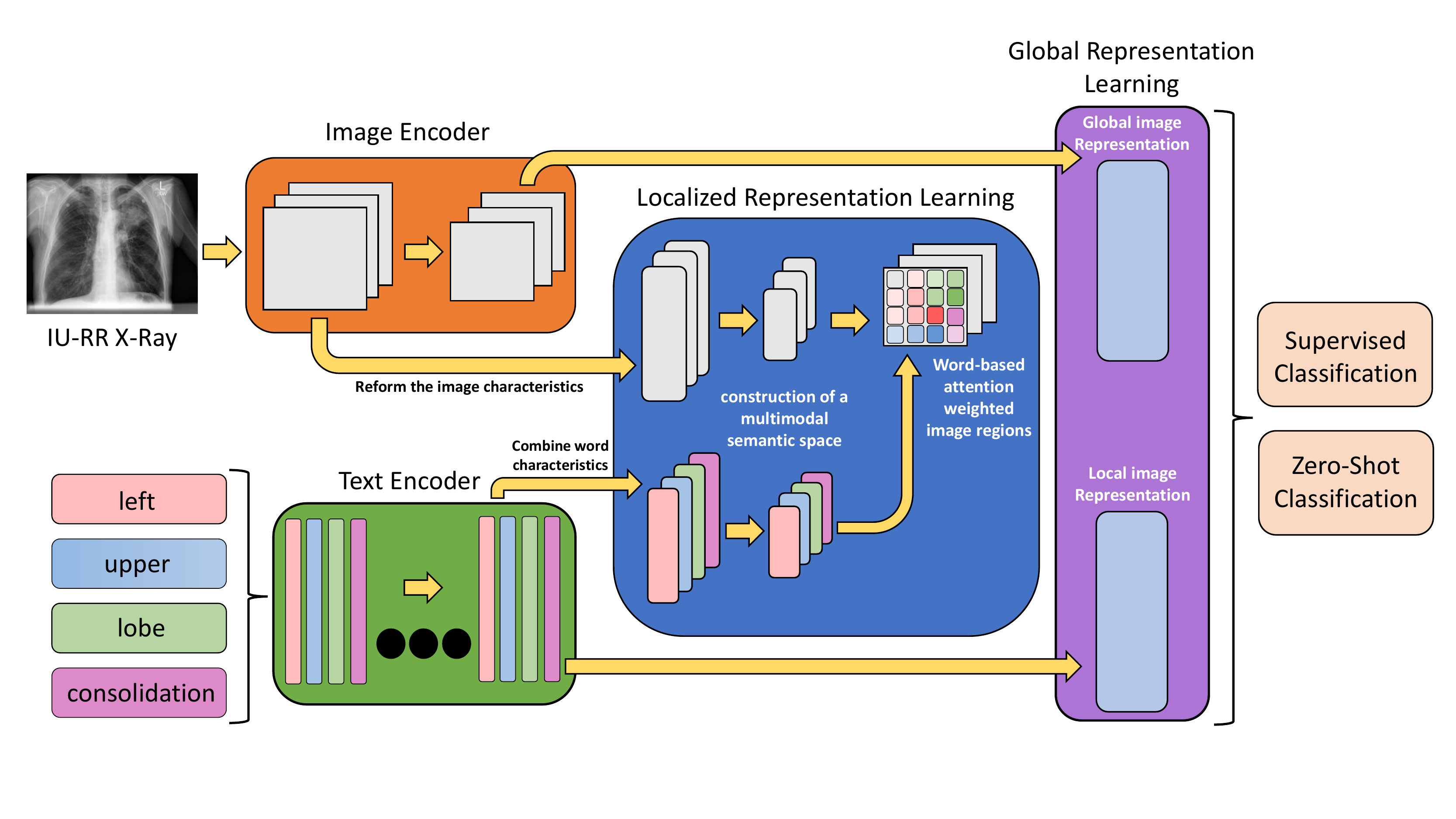}
    \caption{The architecture of multimodal deep representation learning is shown. The model encodes image and text features at global and local levels and then fuses this information to encompass both local and global details for later downstream tasks. The attention mechanism corresponds word-level features to image sub-regions by leveraging a similarity matrix between the local features of an image and its text. Global and local representations are learned through maximizing the posterior probability of the image given text and vice versa and using global and local contrastive learning.}
    \label{fig:architecture}
\end{figure*}

In this paper, we learn multimodal representations of a publicly available dataset of IU-RR and modify and clean the dataset for learning global and sub-regional features of both images and radiology reports. We use a global-local representation framework that attends to both the image-sub-regions and word-piece language and global representations of vision and language modalities. Therefore, as a direct result of this research, radiology reports can be used to attend to related parts in the medical image and learn representations encompassing local and global details of x-ray images and corresponding descriptions. 
The architecture of the method is shown in Fig. \ref{fig:architecture}.

The method here uses a small dataset of Indiana University Radiology Reports (IU-RR) [\citeonline{demner2016preparing}]. Despite its smaller size than other datasets, such as Chexpert, we can train the model and perform downstream tasks such as classification in supervised and zero-shot approaches. 
Our method reaches an area under the curve performance (AUC) of 0.8618, which is slightly higher than the best existing datasets. Finally, we show that we can perform zero-shot classification using multimodal representations.
\newline
This work is embedded in our group’s endeavor to expedite artificial intelligence in biomedical imaging by means of advanced pattern recognition and machine learning methods for computational radiology and radiomics, \textit{e.g.}, [\citeonline{bunte2010adaptive,
8_wismueller2000segmentation,
10_wismuller2004fully,
11_hoole2000analysis,
12_wismuller2006exploratory,
13_wismuller1998neural,
14_wismuller2002deformable,
15_behrends2003segmentation,
16_wismuller1997neural,
17_bunte2010exploratory,
18_wismuller1998deformable,
19_wismuller2009exploration,
22_huber2010classification,
23_wismuller2009exploration,
24_bunte2011neighbor,
25_meyer2004model,
Wismuller2001exploration,
Saalbach2005hyperbolic,
26_wismuller2009computational,
Leinsinger2003volumetric,
27_meyer2003topographic,
Wismuller2009exploration,
29_wismueller2010model,
28_meyer2009small,
meyer2007unsupervised,
31_wismuller2010recent,
meyer2007analysis,
wismuller2000neural,
32_wismueller2008human,
Wismuller2000hierarchical,
wismuller2015method,
33_huber2012texture,
34_wismuller2005cluster,
35_twellmann2004detection,
37_otto2003model,
38_varini2004breast,
Meyer2005computer,
39_huber2011prediction,
40_meyer2004stability,
Wismuller1998hierarchical,
41_meyer2008computer,
Wismuller2013introducing,
45_bhole20143d,
46_nagarajan2013computer,
wismuller2001automatic,
Wismueller1999adaptive,
48_meyer2004computer,
49_nagarajan2014computer,
Pester2013exploring,
50_nagarajan2014classification,
Yang2014improving,
Wismuller2014pair,
Wang2014investigating,
51_wismuller2014framework,
meyer2004local,
Schmidt2014impact,
Nagarajan2015integrating,
Wismuller2015nonlinear,
nagarajan2015characterizing,
Abidin2015volumetric,
Wismuller2016mutual,
Abidin2016investigating,
52_schmidt2016multivariate,
Abidin2017classification,
abidin2017using,
61_dsouza2017exploring,
54_abidin2018alteration,
55_abidin2018deep,
Chockanathan2018resilient,
Dsouza2018mutual,
Chockanathan2019automated,
Abidin2019investigating,
56_dsouza2019classification,
Abidin2020detecting,
Wismuller2020prospective,
Dsouza2020large,
vosoughi2022relation,
Vosoughi2020large,
Vosoughi2021marijuana,
Vosoughi2021schizophrenia,
vosoughi2021_LSAGC,
dsouza2021large,
Vosoughi2021eusipco,
vosoughi2022large}].

\section{Algorithm}

\subsection{Representation Learning Model}
The GLoRIA model [\citeonline{huang2021gloria}] was built on the premise that many anomalies observed in medical imaging tests only occupy a small portion of the image and only roughly match a handful of crucial sentences in medical reports. As a result, the model creates an attention-based framework for multimodal representation learning by contrasting image subregions to the words in the relevant report. This allows for the generation of context-aware local representations of pictures using attention weights that emphasize important visual sub-regions for a particular word. Seen in Fig. \ref{fig:architecture}.

The model takes advantage of the final adaptive average pooling layer of the ResNet-50 model to help with the creation of Image Encoders. The GLoRIA text encoder was pretrained on the BioClinicalBERT model using medical texts from the MIMIC III dataset in order to produce clinically informed text embeddings. Additionally, word-piece tokenization was employed to lessen the embeddings of acronyms and some lexical problems in report drafting. The text encoder extracts feature specific to each word component. The global text feature is the whole of all word-piece features. Through the use of global, semantic data that condenses the picture and report, and local, captures the semantics in the image and, features. Word-level embeddings are used for local text characteristics.
\newline
\textbf{Supervised classification model:} The model trains a linear classifier on top of the already-trained image encoder GLoRIA model to evaluate the data efficiency of the global picture representations.
\newline
\textbf{Zero-shot classification:} Zero-shot Classification analyzes the ability of our learned representations to be categorized without the addition of additional labels. The results of the Zero-shot Classification allow evaluation of how good the representations between reports and images within GLoRIA model are based on the achieved accuracy.
\section{Data}
We used three different datasets in this study due to the lack of radiology reports for CheXpert and CheXphoto, and the smaller size of the IU-RR dataset with radiology reports. In the following, we briefly describe each of these datasets.
\subsection{CheXpert}
\begin{figure*}
   \centering
\begin{tabular}{|l|l|l|l|l|l|}
\hline
Dataset & Atelectasis & Cardiomegaly & Consolidation & Edema & Pleural Effusion \\ \hline
CheXpert [\citeonline{irvin2019chexpert}]&
\includegraphics[width=2.1cm]{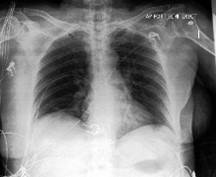}&
\includegraphics[width=2.1cm]{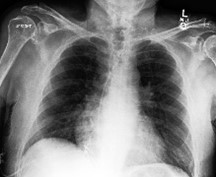}&
\includegraphics[width=2.1cm]{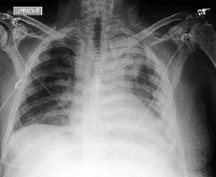}&
\includegraphics[width=2.1cm]{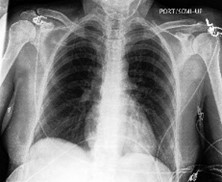}&
\includegraphics[width=2.1cm]{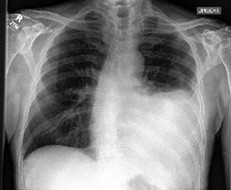}\\ \hline
CheXphoto [\citeonline{phillips2020chexphoto}]&
\includegraphics[width=2.1cm]{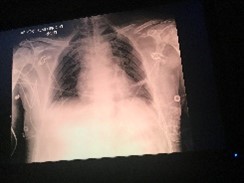}&
\includegraphics[width=2.1cm]{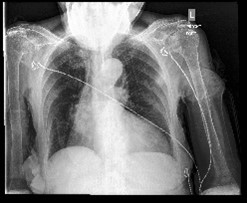}&
\includegraphics[width=2.1cm]{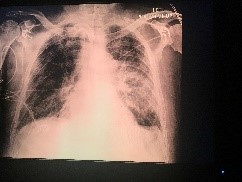}&
\includegraphics[width=2.1cm]{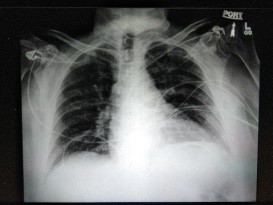}&
\includegraphics[width=2.1cm]{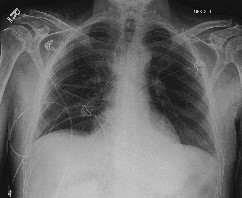}\\ \hline
IU-RR [\citeonline{demner2016preparing}]&
\includegraphics[width=2.1cm]{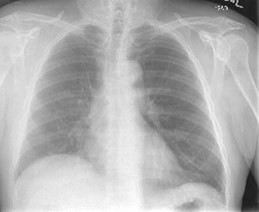}&
\includegraphics[width=2.1cm]{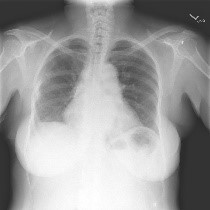}&
\includegraphics[width=2.1cm]{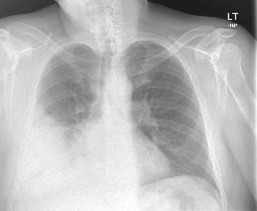}&
\includegraphics[width=2.1cm]{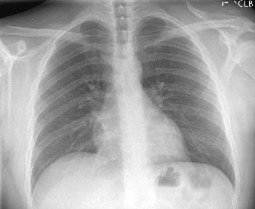}&
\includegraphics[width=2.1cm]{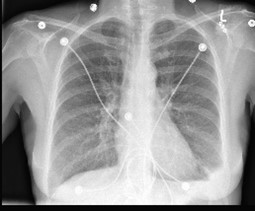} \\ \hline
\end{tabular}
\caption{ Three datasets and their corresponding five types of lung diseases are shown in this image. The radiology reports of CheXpert [\citeonline{irvin2019chexpert}] and CheXphoto [\citeonline{phillips2020chexphoto}] are not publicly available, while IU-RR [\citeonline{demner2016preparing}] are labeled images and their corresponding radiology reports. Despite the smaller size of the IU-RR dataset, we could learn multimodal representations from the IU-RR dataset.}
\label{fig:dataset_examples} 
\end{figure*}

The CheXpert dataset [\citeonline{demner2016preparing}] was used to train and evaluate the representation learning framework for the supervised classification tasks. The 224,316 chest radiographs from the 65,240 individuals in the CheXpert dataset are all linked with the associated radiology reports. A total of 14 medical observations is included on each radiograph's label. This research used 191,229 frontal chest radiograph image-text pairings (original train: 191,027 and original valid: 202). Example of pictures seen in Fig.~\ref{fig:dataset_examples}.

\subsection{CheXphoto}
CheXphoto dataset [\citeonline{irvin2019chexpert}] is a noisy dataset comprised of a training set of natural images and synthetic transformations applied to 10,507 x-rays from 3,000 different patients that were randomly selected from the CheXpert training set. The validation and test sets are composed of 234 x-rays from 200 patients and 668 x-rays from 500 patients, respectively. The GLoRIA model only focuses on the frontal images, so for the CheXphoto, we focus on 28,718 images (original train: 28,112 and original valid: 606). Examples of x-ray images are shown in Fig.~\ref{fig:dataset_examples}.

Both CheXpert and CheXphoto had labels provided for each of the diseases (positive = 1, negative = 0, unclear = -1). Although chest abnormalities are frequently suspected, a conclusive diagnosis cannot be obtained without further testing. However, at the time of this research, the CheXpert radiology reports were still under review for the protected health information review through the Health Insurance Portability and Accountability Act. The radiology reports were not available for both CheXpert and CheXphoto datasets.

\subsection{Chest X-Ray Collection from Indiana University}

The open-access chest X-ray collection from Indiana University (IU-RR) [\citeonline{phillips2020chexphoto}] consists of 3,996 radiology reports from the Indiana Network for Patient Care and 8,121 associated images from the hospitals’ picture archiving systems. The GLoRIA model only focuses on the frontal images so for the Indiana dataset, and we focus on 3,279 images with reports (original train: 2,552 and original valid: 727). The Chexpert-labeler [\citeonline{demner2016preparing}] was used to assign the appropriate labels to the IU-RR images in order to be consistent with the labels provided for CheXpert and CheXphoto.

Multiple alternative labels for a target disease might create conflicting results between categories because our Zero-shot classification and case retrieval are focused on locating the most comparable target. Therefore, only five diseases were selected (atelectasis, cardiomegaly, edema, pleural effusion, and consolidation), and cases with only that disease were extracted from the database [\citeonline{huang2021gloria}].   For CheXpert there are 87 positive images for each of the 5 diseases resulting in CheXphoto\_5x87 dataset (total of 435 images).  For CheXphoto there are 62 positive images for each of the five diseases resulting in the CheXphoto\_5x62 dataset (total of 310 images). Each picture in this collection has a positive annotation for just one particular circumstance. Due to the fact that consolidation was only present in one instance, IU-RR 5x1 could only concentrate on one image per disease (5 cases in total).

The images from the above datasets are split into training, validation, and test sets. For the validation, Chexpert-labeler chooses 5,000 photos randomly from the original expert-labeled training sets for CheXpert and CheXphoto, respectively. For IU-RR Chexpert-labeler chose 500 photos at random from the original training set because of the smaller IU-RR dataset. The training sets for all three datasets are determined by what is left after the corresponding validation sets are selected. Then, from the expert-labeled training file, CheXpert 5x87, CheXphoto 5x62, and IU-RR 5x1 are generated with exclusively positive images. Lastly, the untouched expert-labeled validation file is used as an independent test. Therefore, for CheXpert the frontal train split (CheXpert\_{tr}) consists of 185,778 images, the frontal validation split (CheXpert\_{val}) is 4,301 images, and the frontal test (CheXpert\_{ts}) is 202 images. For CheXphoto the frontal train split (CheXphoto\_{ts}) is 23,512 images, the frontal validation split (CheXphoto\_{val}) is 4,319 images, and the frontal test (CheXphoto\_{ts}) is 606 images. For IU-RR, the frontal train split (IU-RR\_{tr}) is 2,552 images, and the frontal test (IU-RR\_{ts}) is 727 images. The number of images in each split of these datasets is listed in Table~\ref{tab:dataset_splits}.

\begin{table}[t]
\centering
\caption{Three datasets and their corresponding train, validation, and test splits, number of images, and details of each split are listed here.}

\begin{tabular}{|l|l|l|}
\hline
Dataset split      & Description              & Number of images \\ \hline
CheXpert$_{tr}$   & Frontal train split      & 185,778          \\ \hline
CheXpert$_{val}$  & Frontal validation split & 4,301            \\ \hline
CheXpert$_{ts}$   & Frontal test             & 202              \\ \hline
CheXphoto$_{tr}$  & Frontal train split      & 23,512           \\ \hline
CheXphoto$_{val}$ & Frontal validation split & 4,319            \\ \hline
CheXphoto$_{ts}$  & Frontal test             & 606              \\ \hline
IU-RR$_{tr}$      & Frontal train split      & 2,552            \\ \hline
IU-RR$_{ts}$      & Frontal test             & 727              \\ \hline
\end{tabular}

\label{tab:dataset_splits}
\end{table}

\section{Experiments}

\begin{table}[t]
\centering
\caption{Evaluation of our representation learning performance based on supervised learning classification is listed here. The CheXphoto dataset [\citeonline{phillips2020chexphoto}] is used as a noisy dataset to train a classifier based on our representation learning. Both the representation learning dataset and test dataset are from IU-RR [\citeonline{demner2016preparing}]. }
\begin{tabular}{|l|l|l|l|l|l|}
\hline
Train     & Valid     & Test      & Train AUC & Valid AUC & Test AUC \\ \hline
CheXpert$_{tr}$  & CheXpert$_{val}$  & IU-RR$_{ts}$     & 0.680     & 0.681     & 0.856    \\ \hline
CheXpert$_{tr}$  & CheXphoto$_{val}$ & IU-RR$_{ts}$     & 0.676     & 0.644     & 0.865    \\ \hline
CheXpert$_{tr}$  & IU-RR$_{tr}$     & IU-RR$_{ts}$     & 0.674     & 0.881     & 0.868    \\ \hline
CheXphoto$_{tr}$ & CheXphoto$_{val}$ & IU-RR$_{ts}$     & 0.643     & 0.619     & 0.862    \\ \hline
CheXphoto$_{tr}$ & CheXpert$_{val}$  & IU-RR$_{ts}$     & 0.614     & 0.643     & 0.860    \\ \hline
CheXphoto$_{tr}$ & IU-RR$_{tr}$     & IU-RR$_{ts}$     & 0.636     & 0.868     & 0.860    \\ \hline
\end{tabular}\label{tab:supervised_results}
\end{table}

We use the codes from the \url{https://github.com/marshuang80/gloria}, and use their zero-shot classification and supervised classification to evaluate the performance of our representation learning based on the IU-RR dataset that we cleaned and prepared. Area Under the Curve (AUC) is used as the performance metric for these downstream tasks. 

\textbf{Supervised classification}
The goal here is to evaluate the performance of the representation learning on IU-RR dataset based on the performance of the supervised learning problem. To this end, we train the model based on the small-size, publicly available dataset of IU-RR [\citeonline{demner2016preparing}]. Consequently, we use noisy data of CheXphoto to train the classifier based on our encoder. The results are listed  in Table \ref{tab:supervised_results}. As listed in Table \ref{tab:supervised_results}, the performance of the classifier achieves the AUC of 0.8618 $\pm$ 0.0042 (Mean + STD), which is slightly lower than the best of the previously reported values in [\citeonline{xue2019improved}], which is 0.8608. The reason that the test is higher than validation and the train is that the test is on a different data set.

\textbf{Zero-shot classification:} The model here has been applied to three different datasets in various configurations, noting that radiology reports of CheXpert and CheXphoto are not publicly available. Therefore we rely on the IU-RR dataset for the training of the representation learning. However, due to the smaller size of the IU-RR dataset compared to the two other datasets, the zero-shot learning is the model trained on the CheXpert dataset [\citeonline{huang2021gloria}]. 
We compare the performance of the zero-shot classification based on representations that are learned on IU-RR, CheXpert, and CheXphoto, and based on five types of lung diseases, namely atelectasis, cardiomegaly, edema, pleural, and effusion. The detailed values of the classification results are listed in Table \ref{tab:zeroshot_results}.

For comparison, the average AUC for classifying the five lung pathologies on the IU-RR test set ranged from 0.85 to 0.87 using the different training datasets. These results compare favorably to other studies using IU-RR for the classification of the five lung pathologies (AUC range:0.77-0.86) reported in the literature [\citeonline{xue2019improved}].
\begin{table}[]
\centering
\caption{Zero-shot classification performance based on the representation model trained on IU-RR dataset }
\begin{tabular}{|l|l|l|l|l|l|l|}
\hline
  & Atelectasis  & Cardiomegaly  & Consolidation  & Edema  & Pleural Effusion    & Average \\ \hline
CheXpert     & 0.685                & 0.628                 & 0.694                     & 0.754          & 0.717   & 0.696   \\ \hline
CheXphoto    & 0.719                & 0.587                 & 0.700                     & 0.784          & 0.694     & 0.697   \\ \hline
\end{tabular}\label{tab:zeroshot_results}
\end{table}
\section{Conclusion}
We developed a model that uses independent datasets utilizing language and vision modalities to learn a representation encompassing local and global features. The clinical significance of the model is the ability to utilize both image features from the X-Rays and text characteristics from reports to classify lung pathologies on chest X-rays. Five lung pathologies (atelectasis, cardiomegaly, edema, pleural, and consolidation) were considered. The average AUC for classifying the five lung pathologies on the IU-RR test set ranged from 0.85 to 0.87 using the different training datasets. These results compare favorably to other studies using UI-RR to classify the five lung pathologies with a mean AUC of 0.86, which is favorably comparable to existing literature. We have validated the cross-modal model on three independent datasets: IU-RR, CheXpert, and CheXphoto. The robust performance of the model on different datasets is promising and shows that it behaves satisfactorily.

\setstretch{.5}
\tiny
\acknowledgments 
 
This research was partially funded by the American College of Radiology (ACR) Innovation Award “AI-PROBE: A Novel Prospective Randomized Clinical Trial Approach for Investigating the Clinical Usefulness of Artificial Intelligence in Radiology” (PI: Axel Wism{\"u}ller) and an Ernest J. Del Monte Institute for Neuroscience Award from the Harry T. Mangurian Jr. Foundation (PI: Axel Wism{\"u}ller). This work was conducted as a Practice Quality Improvement (PQI) project related to the American Board of Radiology (ABR) Maintenance of Certificate (MOC) for A.W.

\tiny
\setstretch{0.5}
\bibliography{refs, refs_citations} 
\bibliographystyle{spiebib} 

\end{document}